\documentclass[pdflatex]{sn-jnl}

\usepackage{graphicx}%
\usepackage{multirow}%
\usepackage{amsmath,amssymb,amsfonts}%
\usepackage{amsthm}%
\usepackage{mathrsfs}%
\usepackage[title]{appendix}%
\usepackage{xcolor}%
\usepackage{textcomp}%
\usepackage{manyfoot}%
\usepackage{booktabs}%
\usepackage{placeins} 
\usepackage{algorithm}%
\usepackage{algorithmicx}%
\usepackage{algpseudocode}%
\usepackage{listings}%
\usepackage[round,authoryear]{natbib}

\theoremstyle{thmstyleone}%
\theoremstyle{thmstyletwo}%

\theoremstyle{thmstylethree}%

\begin{document}

\title[AI Text Detectors and the Misclassification of Slightly Polished Arabic Text]{AI Text Detectors and the Misclassification of Slightly Polished Arabic Text}


\author[1]{\fnm{Saleh} \sur{Almohaimeed}}\email{salmohaimeed@ksu.edu.sa}

\author[2]{\fnm{Saad} \sur{Almohaimeed}}\email{mohaimeeds@ipa.edu.sa}

\author[1]{\fnm{Mousa} \sur{Jari}}\email{Maljari@ksu.edu.sa}

\author[1]{\fnm{Khaled A.} \sur{Alobaid}}\email{Kalobaid@ksu.edu.sa}

\author[1]{\fnm{Fahad} \sur{Alotaibi}}\email{faalotaibi@ksu.edu.sa}

\affil[1]{\orgdiv{College of Applied Computer Science}, \orgname{King Saud University}, \orgaddress{\city{Riyadh}, \country{Saudi Arabia}}}

\affil[2]{\orgdiv{Department of Digital Transformation}, \orgname{Institution of Public Administration}, \orgaddress{\city{Riyadh}, \country{Saudi Arabia}}}


\abstract{Many AI detection models have been developed to counter the presence of articles created by artificial intelligence (AI). However, if a human-authored article is slightly polished by AI, a shift will occur in the borderline decision of these AI detection models, leading them to consider it as AI-generated article. This misclassification may result in falsely accusing authors of AI plagiarism and harm the credibility of AI detectors. In English, some efforts were made to meet this challenge, but not in Arabic. In this paper, we generated two datasets. The first dataset contains 800 Arabic articles, half AI-generated and half human-authored. We used it to evaluate 14 Large Language models (LLMs) and commercial AI detectors to assess their ability in distinguishing between human-authored and AI-generated articles. The best 8 models were chosen to act as detectors for our primary concern, which is whether they would consider slightly polished human-authored text as AI-generated. The second dataset, Ar-APT, contains 400 Arabic human-authored articles polished by 10 LLMs using 4 polishing settings, totaling 16400 samples. We use it to evaluate the 8 nominated models and determine whether slight polishing will affect their performance. The results reveal that all AI detectors incorrectly attribute a significant number of articles to AI. The best performing LLM, Claude-4 Sonnet, achieved 83.51\%, its performance decreased to 57.63\% for articles slightly polished by LLaMA-3. Whereas the best performing commercial model, originality.AI, achieves 92\% accuracy, dropped to 12\% for articles slightly polished by Mistral or Gemma-3.}

\keywords{Arabic AI Detection, , AI-polished text, Detection fairness}



\maketitle

\section{Introduction}\label{sec1}
\begin{figure}[h]
\centering
\includegraphics[width=0.9\textwidth]{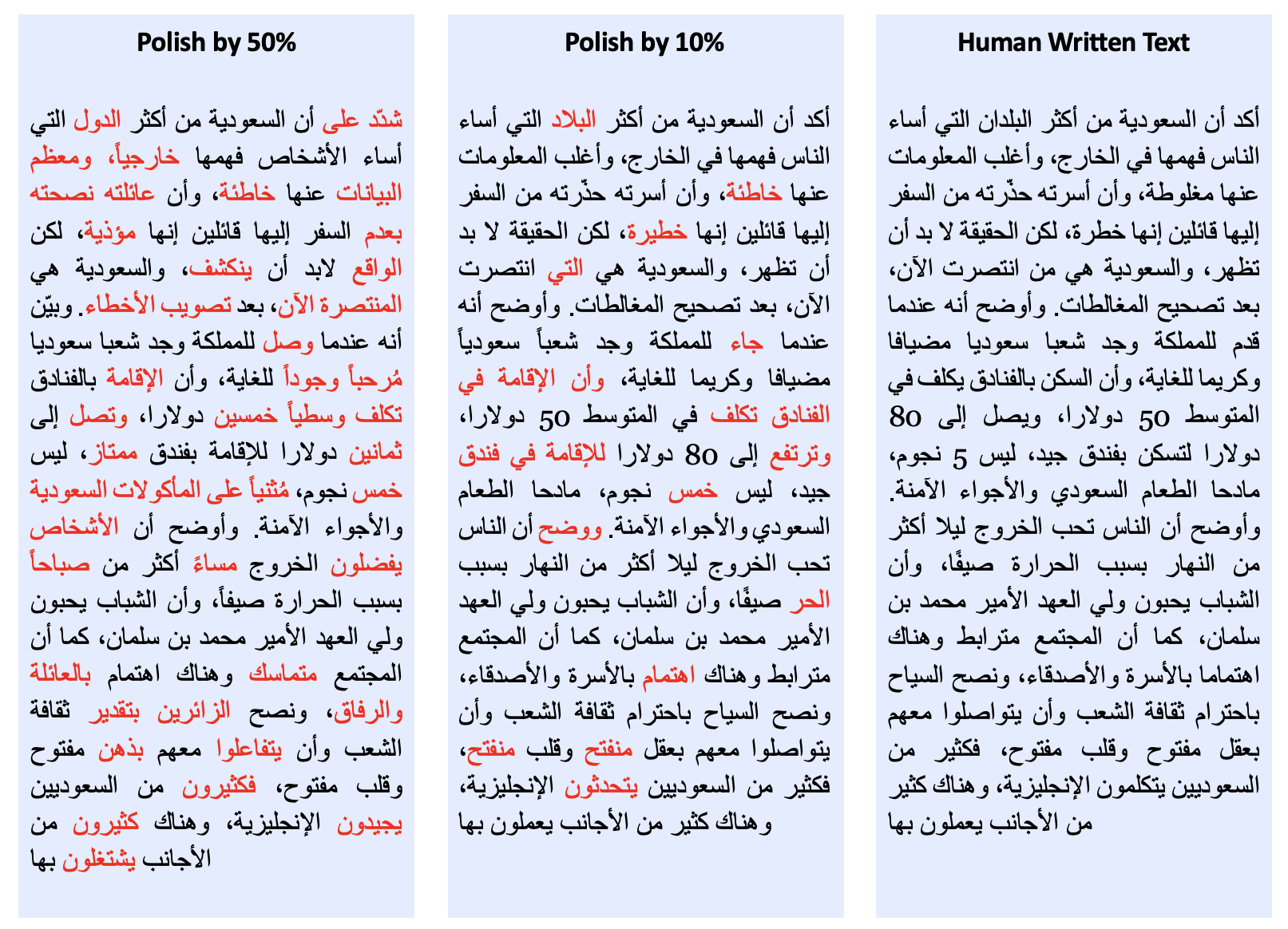}
\caption{This is an example of an Arabic human-written article that has been polished twice under 10\% and 50\% polishing settings. Words with red colors are polished words in comparison to the human-written text}\label{fig1}
\end{figure}

Recent years have seen a great deal of competition between Large Language Model (LLM) companies to provide better multilingual models with an advanced level of capabilities. One of their abilities is to generate text that cannot be distinguished from authentic human writing. Due to this, a large number of English text-detector models have been developed. These detectors are designed to distinguish human-authored text from AI-generated content. However, a critical issue arises when someone uses LLM to slightly polish their own human writing without changing its meaning or adding additional information, as shown in Figure \ref{fig1}. In such cases, text detectors struggle to distinguish whether it is a human-written or entirely generated by artificial intelligence. Consequently, there is a risk of falsely accusing the article of being generated by artificial intelligence when it is not. According to \cite{bib1}, 10.86\% of articles on 246 Fortune 500 companies' blogs are generated by artificial intelligence. However, this may not be true, as authors may only use LLM to enhance the readability of their articles slightly.

From \cite{saha2025}, the authors study this issue and provide clear explanations as to why better models are needed to solve this problem. However, the study focused only on English, and did not give much consideration to other languages, such as Arabic. Compared to many other languages, the Arabic language has a richer morphology and a variety of dialects. Moreover, the use of similar words with different diacritics may lead to a different meaning, even if they are within the same context. Having to deal with these issues makes it more difficult to build a text detector for Arabic articles. 

In this paper, we have conducted a comprehensive investigation using more than 300 experiments in order to address three main concerns. One, what is the effectiveness of popular LLMs and commercial AI models in distinguishing between Arabic human-written articles and those generated by AI. Second, we investigate the effectiveness of these LLMs in terms of polishing text without changing the meaning. Third, we retest the LLMs and commercial AI models to determine whether they consider slightly polished Arabic text as AI-generated.

In order to accomplish this, we first constructed two datasets. The first dataset consists of 800 samples, half of which are human-written and the other half are generated by AI. The second dataset is Ar-APT \footnote{github.com/Saleh-Almohaimeed/Ar-APT} (Arabic AI polish text), which contains 400 authentic Arabic human texts polished by 10 LLMs, each with four different polishing percentage levels, resulting in 16400 samples.

The first dataset was used as a means to address our first concern, which is testing the ability of LLMs and very popular commercial models to differentiate between human-authored articles and AI-generated articles. The second dataset Ar-APT is used to address the aforementioned second and third concerns. We also used Ar-APT to demonstrate our main objective, which is the problem of falsely accusing human-written articles as AI-generated when only slight polishing was done.

There were 14 experiments conducted on the first dataset using different LLMs, such as the GPT-4o \cite{openai2024gpt4o}, LLaMA-3 70B \cite{meta2024LLaMA3}, and Claude-4 Sonnet \cite{anthropic2024claude4sonnet}, as well as commercial models, such as ZeroGPT \cite{zerogpt2023} and Oraginality-AI \cite{originalityai2023}. We will nominate models that achieve acceptable results to act as text detectors on the Ar-APT dataset. After that,  we conducted 320 experiments on the Ar-APT dataset to determine whether nominated LLMs and Commercial models consider human-polished text to be AI-generated or not.

As a result of our findings, it is necessary to design models specifically developed to serve as text detectors for Arabic texts. As the best performing publicly available model is Claude-4 Sonnet \cite{anthropic2024claude4sonnet}, which achieved an accuracy of 83.51\% and a false positive rate of 16.49\%, while the best commercial model achieved 96\% accuracy and 8\% false positive rate.

Moreover, on Arabic slightly polished text, all experimented LLMs and commercial models experienced a drop in performance in determining whether the text was generated using AI or not. Under 10\% polishing percentage, the Claude-4 Sonnet \cite{anthropic2024claude4sonnet} produces an accuracy rate of 65.03\% and a false positive rate of 34.97\% with articles slightly polished by Mistral \cite{mistral2024seba}. Moreover, with articles that were polished by Gemma-3 \cite{google2025gemma3}, Claude-4 Sonnet got 65.71\% accuracy, 34.29\% false positive rate. Thus, with only slightly polished articles, Claude-4 Sonnet's accuracy decreased by approximately 18\%. 

Furthermore, the best performing commercial model, Originality.AI \cite{originalityai2023}, exhibited an impressive performance decline when detecting human articles that were slightly polished (10\%) by Mistral \cite{mistral2024seba} and Gemma-3 \cite{google2025gemma3}. Originality.AI accuracy has dropped to 12\% and the false positive rate jumped to 88\%. This clearly highlights the necessity of addressing this issue, both for researchers and for the owners of the popular Originality.AI \cite{originalityai2023} model.

Moreover, an analysis was conducted to determine whether LLMs are capable of polishing arabic text while keeping the meaning unchanged. In the results, Claude-4 Sonnet performed best, followed by Deepseek-3.1 \cite{deepseek2025v3_1} and GPT-4o \cite{openai2024gpt4o}. Additionally, all of the 10 LLMs used for polishing managed to maintain the meaning and got a high cosine similarity score, except for the Qwen-3 \cite{alibaba2025qwen3} model, which appears to have difficulty with Arabic texts. Contributions:

\begin{enumerate}
\item Constructed the first dataset that contains 800 samples of AI-generated text and human-authored text.
  \item A second dataset (Ar-APT) was constructed using 400 human-authored texts and an additional 16000 AI-polished versions of these texts, generated by 10 LLMs, under 4 degrees of polishing.
  \item 14 experiments were conducted to address our first concern about how well popular LLMs and commercial models distinguish between Arabic human-written articles and AI-generated ones.
  \item Our second concern was addressed by calculating the cosine similarity scores and the Jaccard similarity scores between all polished articles and their original counterparts. In order to determine how effective LLMs are at polishing text without altering the meaning.
  \item 320 experiments were undertaken to answer our third concern, to determine whether LLMs and commercial models consider slightly polished Arabic text to be AI-generated.
\end{enumerate}

\section{Related Works}\label{sec2}
As technology has become increasingly present in all aspects of people's lives. Information is now shared on social media platforms, blog websites, product reviews, and many other places. Consequently, a number of issues have emerged from the misuse of technology. As artificial intelligence tools, particularly deep learning tools, are becoming more prevalent, it has become increasingly important to address these issues and find appropriate solutions. For example, to combat hateful content spread across social media, an NLP task called Hate Speech detection \cite{Almohaimeed2023THOS},\cite{Almohaimeed_2024_ImplicitHate} was designed to identify hateful content. Similarly, text bias detection models \cite{Farrelly2023TopologicalBiasDetection},\cite{Raza2023NBIAS} have been introduced for articles that show bias toward certain races or cultures. An additional instance of technology being misused involves using AI models to generate articles and then falsely claiming that they were written by humans. A new method of addressing this issue has emerged called AI-generated content detection \cite{Chaka2024JALT},\cite{Alshammari2024ArabicAIDetector}, which has led to the development of numerous free and paid models.

In order to finetune these models and distinguish between human-written text and AI-generated text, we need a proper dataset. A number of datasets were released, including M4 \cite{wang2024m4} and RAID \cite{dugan2024raid}.  RAID is not a multilingual dataset, it is only in English. It includes 6 million generated content across 11 models and 8 domains. M4 \cite{wang2024m4} is a noteworthy dataset that spans multiple domains and includes AI-content generated by 8 LLMs. A total of 7 languages were represented in the dataset, including English, Chinese, Urdo, Indonesian, Bulgarian, and Arabic. Additionally, there are several datasets which only support English, such as the MAGE \cite{Li2024MAGE} dataset, and the HC-Var \cite{Xu2024HCVar} dataset. There are also a number of datasets that support a wide variety of languages, such as HC3-Plus \cite{HC3Plus}, and AuText2023 \cite{AuText2023}. In our paper, we built our own dataset for two reasons: in order to ensure diversity, we wanted the Arabic AI Text to be generated by more LLMs, so 10 LLMs were utilized. Secondly, we want to ensure that the human-authored text is sourced from a variety of sources. The majority of datasets focus on news-only resources; however, ours is collected from popular news and blog sources.

Some of the traditional methods used to detect AI-generated articles such as perplexity \cite{Gehrmann2019GLTR}, which emphasize that the LLM generally chooses the highly frequent next word when writing the AI article. Hence, if the next words are mostly too obvious, they flag them as likely to be generated by AI. In a more advanced approach than perplexity \cite{Gehrmann2019GLTR}, models such as BERT \cite{BERTDetect} and XLM-Roberta \cite{MULTITuDE} have been fine-tuned to act as AI detectors. The main purpose of these models is to differentiate between pure artificial intelligence and human-written text. However, if you are including human-written text that has been slightly polished by AI in a way that does not change the meaning, there are no models designed for this purpose. Some studies have addressed the problem without developing AI detecting models, such as \cite{saha2025}. The authors constructed a dataset of 300 samples, then polished the 300 samples using 5 LLMs with 11 polishing settings, producing approximately 15K samples. Each original human-written sample is polished 11 times. They then evaluated 12 detectors and demonstrated that even slight polishing could affect the detector decision, falsely reporting the article was generated entirely by AI. In our paper, we did the same; however, instead of 5 LLMs, we used 10 LLMs for polishing the Arabic article. After we realized that the 11 polishing settings used in their study \cite{saha2025} [1\%, 5\%, 10\%, 20\%, 35\%, 50\%, 75\%, extremely minor, minor, slightly major, and major] did not significantly impact detector performance, we decided to stick to four polishing settings [10\%, 25\%, 50\%, 75\%]. In this way, we generated the same number of samples using more LLMs and fewer polishing settings.

In contrast, very few efforts have been made for the Arabic language, such as \cite{Al-Shaibani2025ArabicAIFingerprint}, which has tested its own dataset of texts extracted from the abstract Arabic research papers and social media reviews. They used 4 LLMs for the generation of AI content, as well as for the detection, including ALLaM \cite{ALLaM}, Jais \cite{Jais}, LLaMA-3.1 \cite{LLaMA3.1}, and GPT-4 \cite{GPT-4}. They also finetuned the XLM-RoBERTa \cite{XLM-R} model to act as detectors, achieving notable results in identifying pure human content from artificial content. However, we did not include their detector in our experiments because according to \cite{originalityai2023}, the model built by Originality.AI performs better and achieves state-of-the-art performance, Therefore, the results from Originality.AI are sufficient for our experiments. Additionally, we are not focusing on comparing every available detector. Instead, we are trying to determine whether slight polishing will affect the decision boundary of leading detectors, and it was very evident from the results of our experiments with Originality.AI \cite{originalityai2023} and ZeroGPT \cite{zerogpt2023}.

\section{Datasets}\label{sec3}
At the beginning, two datasets were constructed by collecting samples from popular news articles and blogs. The sources of the articles are listed in table \ref{tab1}. Three criteria were taken into account. First, we ensure that the articles collected are published in 2020 or before, so that they are ahead of the widespread use of LLMs. Second, to present different types of articles, we have collected articles from eight different domains, each contains 50 articles. Third, approximately half of the dataset was collected as news articles reporting something new or significant. The other half of the articles are simply general articles that discuss information in their field, not necessarily new.

\begin{table}[h]
\caption{Sources of human-written articles used to construct the two datasets}\label{tab1}%
\begin{tabular}{@{}cc@{}}
\toprule
Domain & Sources\\
\midrule
Sport    & \cite{ANAD} \cite{SANAD} \cite{EASC} \cite{site-ijssa} \\
Tourisms    & \cite{ANAD} \cite{SANAD} \cite{site-WikipediaArabic} \cite{site-DollzterTravels2023} \\
Science    &  \cite{EASC} \cite{site-Waqfeya2024} \cite{site-UN2025} \cite{site-WikipediaArabic}\\
Technology    &  \cite{ANAD} \cite{SANAD} \cite{EASC} \cite{site-AAWSAT2025}\\
Business    &  \cite{site-Coins4Arab2025} \cite{EASC}\\
Politics    & \cite{ANAD} \cite{SANAD} \\
Medical    &  \cite{ANAD} \cite{SANAD} \cite{EASC} \\
Finance    &  \cite{SANAD} \cite{site-Coins4Arab2025}\\
\botrule
\end{tabular}
\end{table}

For the first dataset, in addition to 400 articles that were written by humans, 10 LLMs were used to generate 400 additional AI articles. The LLMs that were used were (GPT-4o \cite{openai2024gpt4o}, GPT-3.5 \cite{GPT35Turbo}, LLaMA-3 70B \cite{meta2024LLaMA3}, LLaMA-4 17B \cite{Meta-LLaMA4-17bMaverick}, Deepseek 3.1 \cite{deepseek2025v3_1}, Claude-4 sonnet \cite{anthropic2024claude4sonnet}, Gemma-3 \cite{google2025gemma3}, Qwen-3 \cite{alibaba2025qwen3}, Mistral \cite{mistral2024seba}, and Kimi K2 \cite{KimiK2Instruct} ). Later on, we will use this dataset to assess how well LLMs and commercial AI models can distinguish AI-generated articles from human-generated articles.

\begin{table}[h]
\caption{Comparison Between our Arabic Ar-APT dataset and the english APT dataset}\label{Comparison}%
\begin{tabular}{@{}cccccc@{}}
\toprule
Dataset & Language & \# LLMs & \# Domains & \# Polishing Degrees & \# Samples\\
\midrule
Ar-APT    & Arabic & 10 & 8 & 4 & 16400  \\
APT    & English & 5 & 6 & 11 & 15280  \\
\botrule
\end{tabular}
\end{table}

As for the second dataset, named Ar-APT, it uses the same 400 articles that were written by humans. Following that, we polished the articles at different percentage levels (10\%, 25\%, 50\%, and 75\%), We maintained only those four levels and did not include any more levels due to the fact that in this work \cite{saha2025}, which is similar to our work but done in English. There were 11 levels and results showed that AI-text detectors have difficulty in discriminating between minor and major polishing. As a result, adding a greater degree of AI polishing is not always associated with a higher detection rate. Furthermore, as compared to their work, instead of increasing the number of polishing levels, we have increased the number of LLMs used to polish. They polish their English articles by 5 LLMs, whereas in our work, we polish our Arabic articles by 10 LLMs. Table \ref{Comparison} compares our dataset with theirs. They have added more polishing levels, but we have dominated all other features, which are more crucial to our study. Ar-APT aims to determine whether slightly polished human-written articles will affect the borderline decision of LLMs and commercial AI models with regard to considering these articles to be AI-generated.

Following the collection of the second dataset, we applied two metrics to check whether polishing had been done correctly. For the purpose of determining if human-written text and AI-polished text retained the same meaning, we calculated the cosine similarity score between them. We have also done calculated Jaccard similarly score to verify that some change has actually been made according to the given polishing percentage. In \cite{saha2025} they have removed samples that have a cosine similarity of less than 85\%. However, in our case, the only samples that are excluded are those that are below 5\%. Tables \ref{tab:cosine-similarity} and \ref{tab:jaccard-similarity} illustrates each model along with the level of polishing and the resulting cosine similarity and Jaccard similarity scores.
 
In summary of this section, two datasets were created, one with 800 articles, half of which are human-written and half of which are AI-generated, and another dataset, Ar-APT, with 400 human-written artifices polished by 10 LLMs under four polishing percentages, resulting in a total of 16400 samples.

\section{Experiments}\label{sec4}
\subsection{Experiment setups}\label{subsec1}
When it comes to AI text detectors, there has not been much research conducted in the arabic language. We were looking for models that are available to the general public users and can be used as AI text detectors. Hence, we used the same LLMs used for polishing the articles as AI detectors by only changing the System role prompt to "You are a helpful assistant who always determines if Arabic text is generated by AI or by a Human. Only respond with AI or Human". Additionally, along with the LLMs, we have tested four commercial models, which are Oraginility.AI \cite{originalityai2023}, ZeroGPT \cite{zerogpt2023}, Smodin \cite{SmodinAIDetector2025}, and Isgen \cite{IsgenAIDetector}.

\subsection{Evaluation metrics}\label{subsec2}
There were two metrics used to measure the performance of the detectors: Accuracy, and False Positive Rate (FPR). Accuracy represents the percentage of articles that are not misclassified by the detectors. This is a useful method for determining the general robustness of the model. On the other hand, we use FPR to evaluate the percentage of articles authored by humans that have been misclassified as articles authored by AI. We are more concerned about avoiding falsely accusing the authors of AI plagiarism. Therefore, the FPR will assist us in protecting human authorship and measuring the trustworthiness of the detectors.

\subsection{Experiments Results}\label{subsec3}

In our experiments, we want to answer the following research questions (\textbf{RQs}):
 
\textbf{RQ1}: How effective are current LLMs and commercial AI detectors at detecting articles generated by artificial intelligence?
\textbf{RQ2}: To what extent are LLMs capable of polishing Arabic text while maintaining the meaning?
\textbf{RQ3}: Will polishing the article slightly or extensively affect LLMs decision as to whether the article was created by artificial intelligence?
\textbf{RQ4}: Will polishing the article slightly or extensively affect the decision of the commercial AI detector to determine whether or not this article was generated by artificial intelligence?

\subsubsection{Answering RQ1}

\begin{table}[ht]
\centering
\caption{The accuracy of LLMs and commercial models in detecting AI-generated content. Original  refers to evaluating the models on only human-authored articles. AI refers to evaluating the models on only AI-generated articles. Accuracy measure the overall performance across human-authored and AI-generated articles. FPR measure the model's ability to not label human-authored articles as AI.}
\label{General-performance-of-models}
\begin{tabular}{lcccc}
\toprule
\textbf{Model} & \textbf{Original} & \textbf{AI} & \textbf{Accuracy} & \textbf{FPR} \\
\midrule
\multicolumn{5}{c}{\textbf{Large Language Models (LLMs)}} \\
\midrule
GPT-4 & 99.2 & 15.5 & 57.35 & 0.8 \\
Deepseek 3.1 & 96.87 & 14.0 & 55.44 & 3.13 \\
Mistral & 89.76 & 13.46 & 51.61 & 10.24 \\
Claude-4 Sonnet & 83.51 & 83.17 & 83.34 & 16.49 \\
LLaMA-4 17B & 74.61 & 11.03 & 42.82 & 25.39 \\
Kimi K2 & 74.20 & 75.94 & 75.07 & 25.80 \\
Gemma-3-27B & 52.74 & 63.00 & 57.87 & 47.26 \\
Qwen-3 & 22.50 & 54.00 & 38.25 & 77.50 \\
GPT-3.5 & 6.00 & 79.00 & 42.50 & 94.00 \\
LLaMA-3 70B & 11.00 & 86.25 & 48.63 & 89.00 \\
\midrule
\multicolumn{5}{c}{\textbf{Commercial Detectors}} \\
\midrule
Originality.AI & 92.00 & 100.00 & 96.00 & 8.00 \\
ZeroGPT & 62.00 & 98.00 & 80.00 & 38.00 \\
Isgen & 43.00 & 53.00 & 48.00 & 57.00 \\
Smodin & 20.00 & 90.00 & 55.00 & 80.00 \\
\bottomrule
\end{tabular}
\end{table}

In response to the first question, \textbf{RQ1}, do current LLMs and commercial AI detectors perform well in detecting AI-generated articles? We have tested 10 LLMs and 4 commercial AI detector models. Models that perform well will be nominated for answering \textbf{RQ3} and \textbf{RQ4}. According to table \ref{General-performance-of-models}, there are four different types of metrics. The first metric (Original) measures the performance of the model in identifying human-written (original) text as it is. Similarly, the AI metric shows the performance of the model in classifying AI as AI. Third, (accuracy) measures the overall performance of the model in classifying both original and AI articles. Lastly, (FPR), which measures how often original articles is misclassified as AI.

The nominated LLMs are GPT-4o \cite{openai2024gpt4o}, Deepseek 3.1 \cite{deepseek2025v3_1}, Mistral \cite{mistral2024seba}, Claude-4 Sonnet \cite{anthropic2024claude4sonnet}, kimi K2 \cite{KimiK2Instruct}, LLaMA-4 17B \cite{Meta-LLaMA4-17bMaverick}, and Gemma-3 \cite{google2025gemma3}, while the nominated commercial models are orignality.AI \cite{originalityai2023} and ZeroGPT \cite{zerogpt2023}. We chose these models for the following reasons. The FPR of GPT-4o, Deepseek 3.1, and Mistral is 6\%, 3.13\%, and 10.24\%, respectively. This provides a greater level of assurance that human author articles will not be classified as AI. Meanwhile, they unfortunately classify the majority of AI as human-written (original), achieving an overall accuracy of 57.35\%, 55.44\%, and 51.61\%, respectively. That is to say, they should be evolved in this regard in order to be used as AI detectors. The 3 LLMs strictly adhere to the principle of not falsely accusing original articles as AI-generated. Therefore, these LLMs are being nominated to serve as AI detectors to determine whether minor or major polishing will affect their decision and increase the FPR percentage, or whether they will continue to adhere to not falsely labeling original articles as AI.

In the case of Claude-4 Sonnet, which received 16.49\% FPR, and Kimi K2, which received 25.8\% FPR. Both models demonstrate good performance and acceptable FPR. Additionally, Claude-4 Sonnet and Kimi K2 achieved the highest overall accuracy, with 83.34\% percent and 75.07\%, respectively. These results indicate that these models are the most accurate among the 10 LLMs. Specifically, Claude-4 Sonnet achieved the highest level of accuracy as well as a low FPR. Thus, we have nominated them to act as AI detectors in the following experiments. The last two candidates to act as AI detectors in the following experiments are LLaMA-4 17B and Gemma-3. We consider these two even though they do not have a very low FPR as GPT-4o, Deepseek 3.1, and Mistral, and have not achieved a good performance compared to Claude-4 Sonnet and Kimi K2, they are included because they achieve an FPR lower than 50\%. Including them will allow us to compare their results to those of the previous LLMs, in order to determine if polishing affects all models equally or if it has a more significant impact on models with lower performance. The three last models are Qwen-3, GPT 3.5, and LLaMA-3 70B. Achieve high FPRs of 77.5\%, 94\%, and 89\%, respectively. Since they mostly misclassify human-authored articles as AI, they cannot be trusted to act as AI detectors. Therefore, these three models have not been nominated for the second set of experiments.

Regarding commercial AI detectors models. It was important to include them because many companies and educational sectors rely on them to verify the authenticity of their articles. In addition, previous LLMs did not develop with the intention of serving as AI article detectors, whereas these commercial AI models do. There are two things we want to make sure: that they are trustworthy and that polishing will not affect their decisions. The first nominated Originality.AI \cite{originalityai2023} has been selected as it obtained an impressive 8\% FPR and 96\% overall accuracy, which makes it the best model and very useful for our subsequent experiments to determine whether or not slightly polishing affects their results. ZeroGPT \cite{zerogpt2023} is the second nominated product, with a FPR of 38\% and an overall accuracy of 80\%. We have not nominated the last two sites, Smodin \cite{SmodinAIDetector2025} and Isgen \cite{IsgenAIDetector}, as their results were not sufficient for the following experiments.

Additionally, our data contains 400 human-authored articles. It is important to note that when testing the performance of commercial models, we only test 100 articles and their polished versions. It is due to the fact that we were not able to obtain an API key in order to automate the process of testing articles. The process is carried out manually one by one. The two nominated models were tested on 100 original articles, and since each article was then polished 40 times with 10 models and under 4 different polishing settings, A total of 8000 articles were manually tested. Due to this, we did not test the entire dataset since we would need to manually test 32000 articles. Further, our main objective is to demonstrate that these commercial AI models will be affected by slight polishing, and our analysis of 100 articles is sufficient to demonstrate that.

\subsubsection{Answering RQ2}

\begin{table}[h]
\centering
\caption{Average cosine similarity scores between original (human-authored) and AI-polished articles at different polishing levels}
\label{tab:cosine-similarity}
\begin{tabular}{lcccc}
\toprule
\textbf{Model} & \textbf{10\%} & \textbf{25\%} & \textbf{50\%} & \textbf{75\%} \\
\midrule
GPT-4o & 94.16 & 93.14 & 92.92 & 91.26 \\
Deepseek 3.1 & 94.65 & 93.58 & 92.50 & 90.34 \\
Mistral & 94.82 & 94.44 & 92.70 & 92.05 \\
Claude-4 Sonnet & 95.75 & 94.62 & 89.34 & 89.16 \\
LLaMA-4 & 93.28 & 91.30 & 90.51 & 88.20 \\
Gemma-3 & 93.12 & 93.15 & 92.54 & 90.54 \\
Qwen-3 & 80.94 & 82.20 & 81.66 & 81.46 \\
GPT-3.5 & 92.66 & 92.63 & 92.67 & 92.28 \\
LLaMA-3 70B & 92.25 & 88.95 & 88.06 & 85.40 \\
Kimi K2 & 92.26 & 88.28 & 89.39 & 88.36 \\
\bottomrule
\end{tabular}
\end{table}

\begin{table}[h]
\centering
\caption{Average Jaccard similarity scores between original (human-authored) and AI-polished articles at different polishing levels}
\label{tab:jaccard-similarity}
\begin{tabular}{lcccc}
\toprule
\textbf{Model} & \textbf{10\%} & \textbf{25\%} & \textbf{50\%} & \textbf{75\%} \\
\midrule
GPT-4 & 66.68 & 47.50 & 41.21 & 29.01 \\
DeepSeek (DEEPS) & 70.03 & 58.71 & 45.39 & 28.50 \\
Mistral (MIST) & 64.85 & 50.67 & 45.17 & 36.81 \\
Claude-4 & 89.15 & 80.00 & 35.93 & 34.82 \\
LLaMA-4 & 64.90 & 54.79 & 45.01 & 35.17 \\
Gemma-3-27B & 45.38 & 41.15 & 40.63 & 26.40 \\
Qwen & 36.45 & 27.73 & 30.82 & 19.75 \\
GPT-3.5 & 73.81 & 73.59 & 73.72 & 73.62 \\
LLaMA-3-70B & 60.96 & 48.48 & 42.99 & 32.55 \\
Kimi & 32.22 & 23.24 & 23.97 & 20.81 \\
\bottomrule
\end{tabular}
\end{table}

The table \ref{tab:cosine-similarity} shows that LLMs polish the data and retain the meaning to such an extent that all LLMs, with the exception of Qwen-3, \cite{alibaba2025qwen3}, achieve an average cosine similarity score above 92\% under 10\% polishing settings. Increasing the polishing percentage reduces this cosine similarity, but not by a substantial amount. LLaMA-3 70B is the only model that decreases dramatically, \cite{meta2024LLaMA3}, achieving a good cosine similarity of 92.25\% at a polishing setting of 10\% and drops to 85.4\% at a polishing setting of 75\%.
 
The best model for preserving the meaning between them is Mistral \cite{mistral2024seba}, which achieves an average of 93.5\% across all polishing settings. Then followed by GPT-4o \cite{openai2024gpt4o}, and Deepseek 3.1 \cite{deepseek2025v3_1}, each of which achieved an average of 92.87\% and 92.77\% respectively. The worst model is Qwen-3, which achieves an average of 81.57\%. This indicates that Qwen-3 is not suitable for polishing Arabic texts.
 
A cosine similarity score of less than 100\% between two sentences does not necessarily mean that they do not have the exact meaning; rather, it indicates that in the embedding space, the sentences do not share the exact placement of words. For example, the two sentences “He began working this morning.” and “He began working today morning.”. These sentences have exactly the same meaning in the reader's mind, but the cosine similarity score is 96.49\% with SBERT \cite{SentenceBERT} embedding space and 97.64\% with MPNet \cite{MPNet} embedding space.  Accordingly, a polished article having more than 90\% cosine similarity score indicates that it preserves the original meaning. All used LLMs except Qwen-3 are considered to be good models for polishing Arabic text while maintaining its meaning.

From table \ref{tab:jaccard-similarity}, we can see similar ideas to table \ref{tab:cosine-similarity}. The Jaccard similarity score is highest at the 10\% polishing setting and decreases as the polishing percentage is increased. Moreover, we notice that it does not adhere strictly to the percentage of polishing given to them. At a polishing setting of 10\%. The best model is Claude-4 Sonnet \cite{anthropic2024claude4sonnet}, which achieved 89.15\%. After that, GPT 3.5 \cite{GPT35Turbo}, Deepseek 3.1 \cite{deepseek2025v3_1}, and GPT-4o \cite{openai2024gpt4o} respectively achieved 73.81\%, 70.03\%, and 66.68\%. It is acceptable to consider this as a minor change as long as the Jaccard similarity score is above 60\% in our experiment. Therefore, Qwen-3 \cite{alibaba2025qwen3}, Kimi K2 \cite{KimiK2Instruct} and Gemma-3 \cite{google2025gemma3} cannot be considered as having done minor polishing under 10\% and 25\% polishing settings.
 
To demonstrate how the Jaccard similarity calculation is performed, we will use these two sentences, the first sentence of 10 words. "we went for the grocery store to buy some milk". The second sentence also contains 10 words. "we went for the grocery store to get some milk". Despite the fact that they both contain exactly the same number of words and the same words with one exception. Since the Jaccard calculation formula is (Total intersect words/Total unique words in both sentences), the Jaccard similarity score between these two sentences will be 81.82\%. As a measure of polishing quality, it is helpful to use Jaccard to verify that the text has not undergone too many changes.
 
As a final conclusion from these two tables. Answering \textbf{RQ2}, all 10 LLMs are capable of polishing arabic text, but their performance differs from one another, with Claude-4 Sonnet being the best followed by Deepseek 3.1 and GPT-4o being the next-best. Therefore, for \textbf{RQ3} experiments, we should pay more attention to Claude-4 Sonnet results. since it attains a high cosine similarity score and follows a good polishing percentage. Afterwards, Deepseek 3.1 and GPT-4o both achieved good results, but were not on the same level as Claude-4 Sonnet. Follow that Mistral \cite{mistral2024seba}, LLaMA-3 70B \cite{meta2024LLaMA3} and LLaMA-4 17B \cite{Meta-LLaMA4-17bMaverick}. They all achieved good results, but not at the same level as the above-mentioned models. The cosine similarity and Jaccard similarity scores for GPT-3.5 \cite{GPT35Turbo} were similar in both tables, indicating that it does not listen to the prompt instructions. The situation of GPT-3.5 has been reported in several publications such as \cite{Almohaimeed2024GAT-SQL} that it is not that good at following Arabic instructions. The Jaccard similarity scores of Qwen-3, Kimi K2, and Gemma-3 are very low, under 10\% and 25\% polishing settings. Thus, when answering \textbf{RQ3}, if these three models consider the article to be AI-generated, it will not be taken into account.

\subsubsection{Answering RQ3}
\begin{figure}[h!]
\centering

\begin{minipage}[b]{0.49\textwidth}
    \centering
    \includegraphics[width=\textwidth]{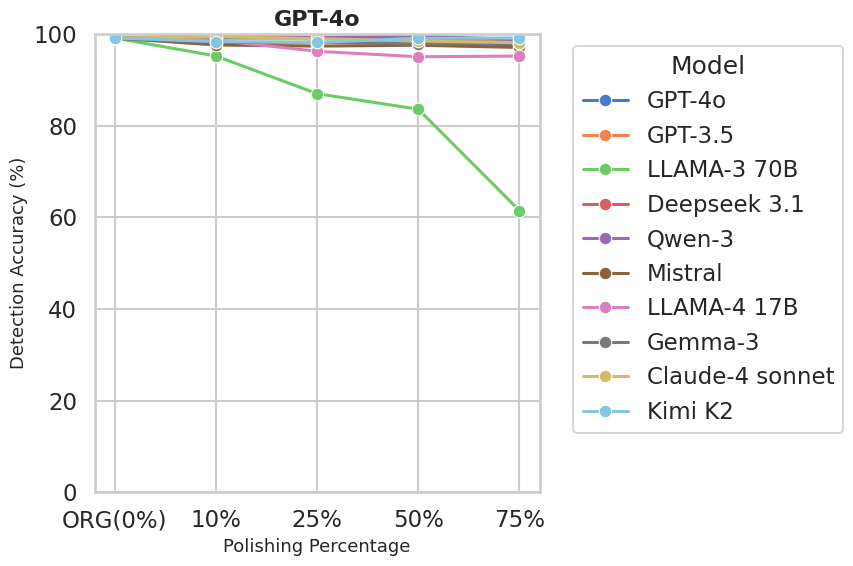}
    \caption{AI-text detection rate for GPT-4o with 5 different polishing settings.}
    \label{fig:1a}
\end{minipage}
\hfill
\begin{minipage}[b]{0.49\textwidth}
    \centering
    \includegraphics[width=\textwidth]{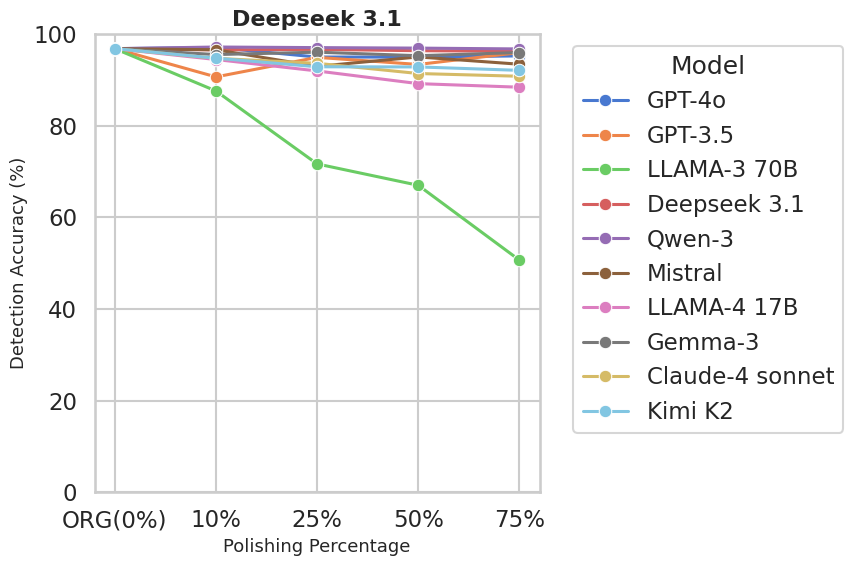}
    \caption{AI-text detection rate for Deepseek 3.1 with 5 different polishing settings.}
    \label{fig:1b}
\end{minipage}

\vspace{0.4cm}

\begin{minipage}[b]{0.49\textwidth}
    \centering
    \includegraphics[width=\textwidth]{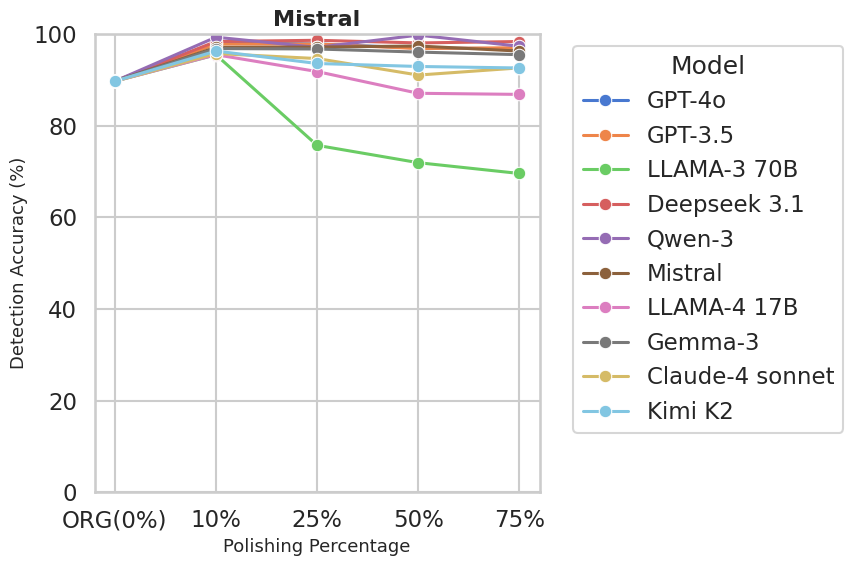}
    \caption{AI-text detection rate for Mistral with 5 different polishing settings.}
    \label{fig:1c}
\end{minipage}
\hfill
\begin{minipage}[b]{0.49\textwidth}
    \centering
    \includegraphics[width=\textwidth]{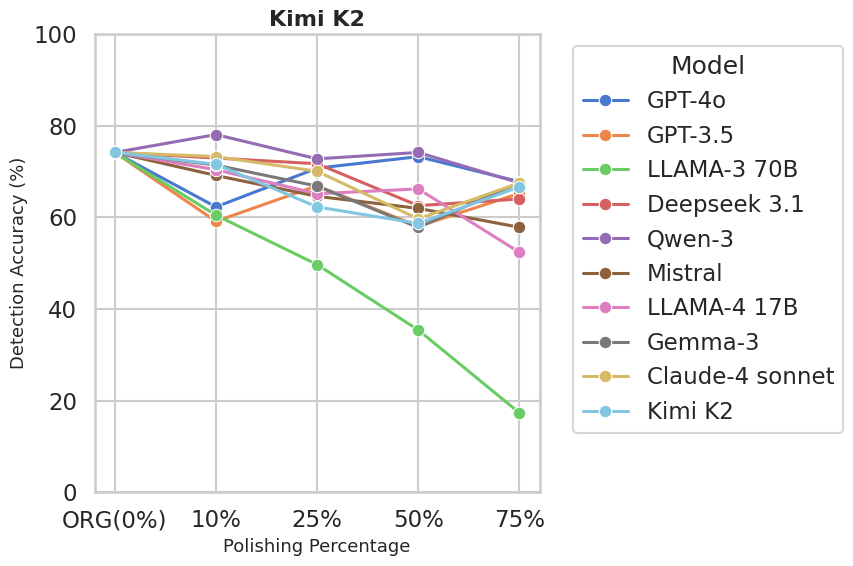}
    \caption{AI-text detection rate for Kimi K2 with 5 different polishing settings.}
    \label{fig:1d}
\end{minipage}

\vspace{0.4cm}

\begin{minipage}[b]{0.49\textwidth}
    \centering
    \includegraphics[width=\textwidth]{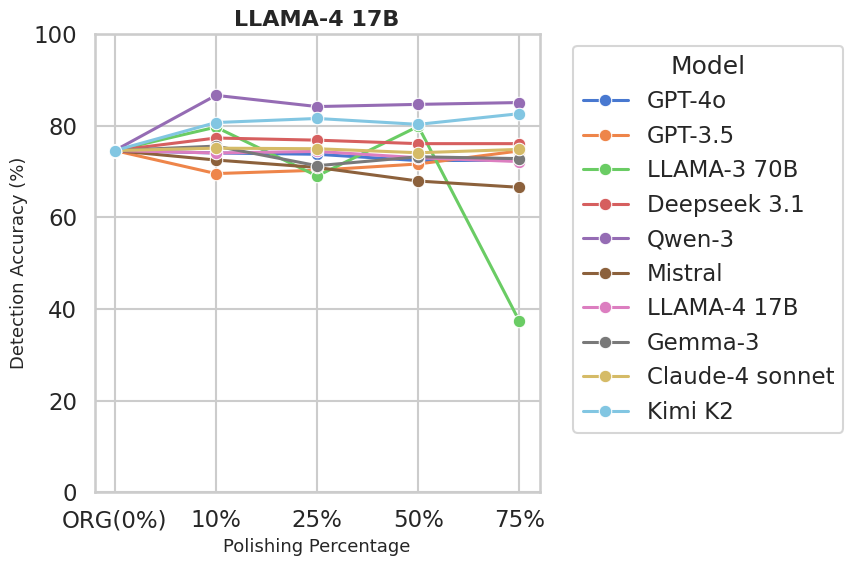}
    \caption{AI-text detection rate for LLaMA-4 17B with 5 different polishing settings.}
    \label{fig:1e}
\end{minipage}
\hfill
\begin{minipage}[b]{0.49\textwidth}
    \centering
    \includegraphics[width=\textwidth]{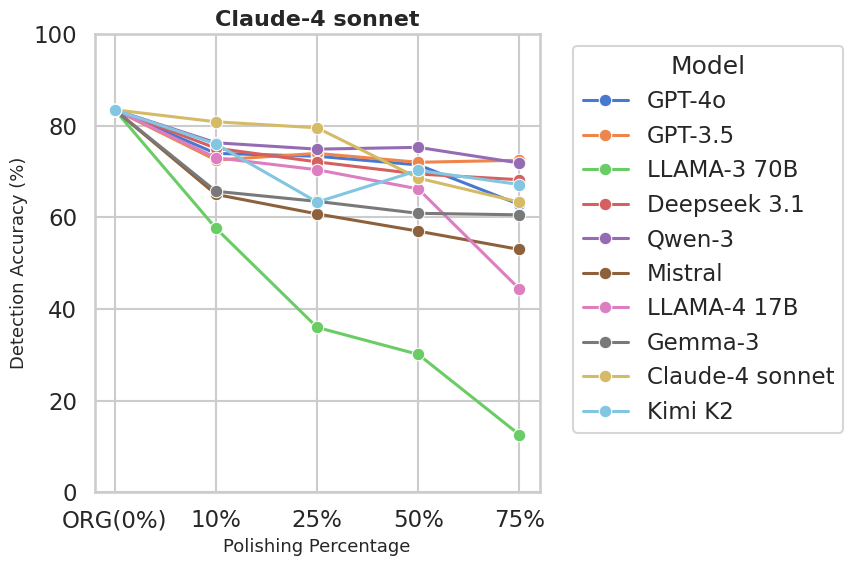}
    \caption{AI-text detection rate for Claude-4 Sonnet with 5 different polishing settings.}
    \label{fig:1f}
\end{minipage}

\end{figure}

In this experiment, our goal is to demonstrate that we do not want to falsely accuse writers that their articles are AI-generated. As they only use current LLMs to improve their own writing without changing the meaning or changing too much of their own style. Detailed results of the experiments are presented in figures [\ref{fig:1a},\ref{fig:1b},\ref{fig:1c},\ref{fig:1d},\ref{fig:1e},\ref{fig:1f}] and as numerical representations in tables [\ref{10P-polishing}, \ref{25P-polishing}, \ref{50P-polishing} \ref{75P-polishing}] in Appendix \ref{AP1}.

From our answer to \textbf{RQ1}, we have stated that the GPT-4o, Deepseek 3.1, and Mistral models strictly adhere to the principle of not falsely accusing original articles of being generated by artificial intelligence. In the case of GPT-4o \cite{openai2024gpt4o} detectors, which achieved 99.2\% under 0\% polishing, the performance of detecting did not significantly change under 10\% polishing, with the exception of articles polished by LLaMA-3 70B, which achieved 95.26\%. Approximately 5\% of articles are falsely accused of being AI-generated. Under 25\% polishing setting, detecting articles polished by Mistral is 97.30\%, LLaMA-4 17B is 96.30\%, and LLaMA-3 70B is 87.04\%. This illustrates that even a robust model such as GPT-4o can become affected by minor or mid-minor polishing. Additionally, GPT-4o performed well and did not accuse other articles polished by Deepseek 3.1, Qwen-3, and Claude-4 Sonnet, having all achieved greater than 99\% under 10\% and 25\% polishing settings. The OpenAI \cite{OpenAIPlatform} platform offers researchers a way to partially finetune their models. Future researchers may utilize this functionality to partially fine-tune GPT-4o and make it differentiate between AI-generated and slightly polished human-authored articles.

Deepseek 3.1 \cite{deepseek2025v3_1}, which received 3.13 FPR, was expected to not be affected too much by slightly polishing articles, but that was not the case. When the articles were polished by LLaMA-3 70B, deepseek 3.1 obtained a 9.24\% lower percentage under 10\% polishing settings and a 24.91\% lower percentage under 25\% polishing settings. Even though it was not affected by the polishing of other models as happened with LLaMA-3 70B, it is a red flag that we need to consider in order to avoid false accusations. The same problem also occurs with Mistral \cite{mistral2024seba}, which performs well across all articles generated by other models except LLaMA-3 70B. There is no doubt that the LLaMA-3 70B polishing style structure tricks the model into thinking that it is AI-generated. Even under 50\% and 75\% polishing settings, GPT-4o, Deepseek 3.1 and Mistral decrease slightly in all other nine models with the exception of LLaMA-3 70B. This illustrates that LLaMA-3 70B approach in handling Arabic words differs from all other models. The effect of LLaMA-4 17B is somewhat similar to that of LLaMA-3 70B, however, it is not by as much.

Furthermore, The best performing LLM, Claude-4 Sonnet \cite{anthropic2024claude4sonnet}, when evaluated under 10\% polishing settings, we can observe that it has become the most affected LLM. In the case of 0\% polishing, the level of accuracy was 83.51\%. Under 10\% polishing, the level of accuracy decreased significantly for all LLMs, where the lowest decrease was obtained with articles polished by their own, with a decrease of 3\%, and the highest decrease was obtained with articles polished by LLaMA-3 70B, achieving approximately 26\% lower accuracy. Under 25\% polishing, results decreased by more than 20\% with models such as Kimi K2, Gemma-3, and Mistral. Additionally, performance of Claude-4 Sonnet with articles polished LLaMA-3 70B experienced a significant reduction by almost 50\%. The significant reduction in accuracy emphasizes the need for building an AI detector that can distinguish between articles created by AI and articles that have been slightly polished by humans.

LLaMA-4 17B \cite{Meta-LLaMA4-17bMaverick} and Kimi K2 \cite{KimiK2Instruct} did not achieve that good results. Their accuracy was 75.07\% and 42.82\%, respectively, and their FPR was 25.8\% and 25.39\%. The purpose of reporting the results of these two detectors is to illustrate the difference between their performance and that of better LLM detectors, including GPT-4o, Deepseek 3.1, Mistral, and Claude-4 Sonnet. Both of them obtained similar results. However, Kimi K2 has a lower FPR rate when increasing the polishing percentage, but this is not surprising given that the model is not constructed to function as an AI detector.

\subsubsection{Answering RQ4}

\begin{table}[h]
\centering
\caption{AI-text detection accuracy for Originality.AI commercial model under 4 different polishing settings, Model refers to the ones used for polished the text}
\label{Originality results}
\begin{tabular}{@{}lcccc@{}}
\toprule
\textbf{Model} & \textbf{10\%} & \textbf{25\%} & \textbf{50\%} & \textbf{75\%} \\
\midrule
GPT-4o        & 23 & 23 & 12 & 0 \\
GPT-3.5     & 74 & 74 & 74 & 74 \\
Kimi K2       & 28 & 28 & 25 & 35 \\
Claude-4 Sonnet      & 90 & 90 & 74 & 71 \\
LLaMA-3 70B   & 76 & 76 & 46 & 44 \\
deepseek 3.1    & 60 & 60 & 32 & 21 \\
Qwen-3        & 18 & 18 & 18 & 0 \\
Mistral        & 12 & 12 & 9  & 7  \\
LLaMA-4 17B   & 71 & 71 & 37 & 35 \\
Gemma-3       & 12 & 12 & 12 & 5  \\
\botrule
\end{tabular}
\end{table}

\begin{table}[h]
\centering
\caption{AI-text detection accuracy for ZeroGPT commercial model under 4 different polishing settings, Model refers to the ones used for polished the text}
\label{ZeroGPT results}
\begin{tabular}{@{}lcccc@{}}
\toprule
\textbf{Model} & \textbf{10\%} & \textbf{25\%} & \textbf{50\%} & \textbf{75\%} \\
\midrule
GPT-4o        & 33 & 33 & 31 & 29 \\
GPT-3.5     & 53 & 51 & 54 & 53 \\
Kimi K2        & 50 & 48 & 47 & 45 \\
Claude-4 Sonnet      & 57 & 57 & 57 & 57 \\
LLaMA-3 70B   & 55 & 53 & 53 & 57 \\
deepseek 3.1    & 54 & 53 & 51 & 50 \\
Qwen-3        & 42 & 42 & 42 & 41 \\
Mistral        & 31 & 31 & 31 & 31 \\
LLaMA-4 17B   & 43 & 43 & 43 & 47 \\
Gemma-3       & 34 & 34 & 33 & 33 \\
\botrule
\end{tabular}
\end{table}

According to table \ref{General-performance-of-models}, Originality.AI \cite{originalityai2023} has achieved a total accuracy of 96\% and a false positive rate of 8\%. When it is tested under four different polishing settings. We can observe how badly it is affected under even the smallest amount of change. The performance of Originality.AI on articles polished by GPT-4o decreases by 69\%. It is likely that these models are built to prove themselves to be effective AI detectors, and since GPT-4o is currently the most commonly used model. The Originality.AI tries to demonstrate its abilities by training on large amounts of data collected from such a model. However, for models like Claude-4 Sonnet under 10\% and 25\% polishing settings, the results of the Originality.AI detection are not dramatically different. This is likely because the data used to train the Originality.AI model does not have the AI article style generated by Claude-4 Sonnet. Moreover, for articles polished using Mistral or Gemma-3, the Originality.AI model falsely accuses all articles except for 12\%.

This calls for researchers and commercial AI developers to construct models that can differentiate between AI-generated articles and original articles that have been slightly polished by AI. The problem of falsely accusing authors of utilizing an AI-generated tool to generate their text is more significant than failing to address actual violators. As for the results related to the ZeroGPT \cite{zerogpt2023}, even though it did not achieve very high results in Arabic as Originality.AI, ZeroGPT achieved 80\% total accuracy and 38\% FPR. Under 10\% polishing settings, the highest decrease in results was for articles polished by Mistral, GPT-4o, and Gemma-3, which amounted to 31\%, 33\%, and 34\%, respectively. 

In table \ref{Originality results}, We can observe that GPT-4o and Claude-4 Sonnet experiences a sharp decline under the 4 polishing setting. This is due to the level of Jaccard similarity differences observed in GPT-4o and Claude-4 Sonnet in table \ref{tab:jaccard-similarity}. As opposed to GPT 3.5, which is very stable across all polishing settings. It is due to the same reason that the Jaccard and cosine similarities scores remained the same under all four polishing settings in tables \ref{tab:cosine-similarity} and \ref{tab:jaccard-similarity}. For results related to ZeroGPT in table \ref{ZeroGPT results}, In the case of Claude-4 Sonnet, we observe the same results 57\% regardless of the polishing setting, even though the Jaccard similarity varies among the polishing settings. It is probably because models such as ZeroGPT are trained to detect articles generated by GPT-4o and LLaMA-3 as they stated in their official website, but did not pay much attention to models such as Claude-4 Sonnet, which are less commonly used. We have also noticed that for many models, such as the LLaMA-3 70B, Deepseek 3.1, Qwen-3, there is no significant difference when polishing under 10\% or 75\%. This could be for the same reason mentioned with articles polished by Claude-4 Sonnet.

\section{Conclusion}\label{sec5}

This paper presents a comprehensive analysis of the performance of 10 popular large language models such as GPT-4o, LLaMA-3, and Claude-4 Sonnet, as well as 4 commercial AI detection such as Originality.AI and ZeroGPT, in order to address two primary concerns. First concern focuses on evaluating the ability of those models to distinguish between Arabic articles authored by humans and Arabic articles generated by AI. To assess this concern, we have constructed a dataset consisting of 800 samples. The best model results were obtained with the Claude-4 Sonnets, which had an overall accuracy of 83.51\% and a FPR of 16.49\%. Additionally, Originality.AI is the best commercial AI model, with an accuracy rate of 96\% and a FPR of 8\%. On the other side, our second concern focuses on evaluating the same models to determine if using AI to slightly polish a human-written article without changing its meaning will result in changing the decision of detectors from a human-written article to an entirely AI-generated article. It is intended to assess the sensitivity of these detectors to slight polishing applied to human-written text. An Ar-APT dataset of 16400 samples has been constructed to assess this concern, and the results indicate that the best performing LLM, Claude-4 Sonnet, was adversely affected across all articles polished by any of the 10 LLMs. Articles generated by LLaMA-3 70B demonstrated the highest performance decrease under 10\% polishing settings, with Claude-4 Sonnet performance decreasing from 83.51\% to 57.63\%. On the other side, Commercial AI models were more affected than LLMs as the performance of Originality.AI models declined by 80\% from 92\% to 12\% with articles polished by Mistral or Gemma-3 under 10\% polishing setting. To avoid falsely accusing authors of AI palagrisms and to increase the trustworthiness of AI detectors, researchers and AI detector developers should pay more attention to consider slightly polished articles as not AI generated.

\vspace{1cm}

\noindent \textbf{Acknowledgments} This research was funded by the Ongoing Research Funding program, (ORF-2025-XXX), King Saud University, Riyadh, Saudi Arabia.

\vspace{1cm}

\noindent \textbf{Author contributions:} Saleh Almohaimeed: Conceptualization, Study Design, Dataset Preparation, Experimentation, Result Analysis, Writing Original Draft, Writing Review and Editing.
Saad Almohaimeed: Conceptualization, Comparative Experiments, Model Evaluation, Writing Review and Editing.
Mousa Jari: Conceptualization, Writing Review and Editing.
Khaled A. Alobaid: Conceptualization, Writing Review and Editing.
Fahad Alotaibi: Conceptualization, Writing—Review and Editing.

\vspace{1cm}

\noindent \textbf{Data availability:} The data in this study are available at https://github.com/Saleh-Almohaimeed/Ar-APT

\vspace{1cm}

\noindent This is a preprint version of the manuscript.

\bibliographystyle{plainnat}
\bibliography{sn-bibliography}

@article{bib1,
  author    = "Gillham, Jonathan",
  title     = "Study Finds Almost 11% of Fortune 500 Blog Articles Are Likely AI-Generated",
  journal   = "Originality.ai Blog",
  volume    = "",
  number    = "",
  pages     = "",
  year      = "2025"
}

@article{saha2025,
  author    = "Saha, Shoumik and Feizi, Soheil",
  title     = "Almost AI, Almost Human: The Challenge of Detecting AI-Polished Writing",
  journal   = "Findings of the Association for Computational Linguistics: ACL 2025",
  volume    = "",
  number    = "",
  pages     = "25414--25431",
  year      = "2025"
}

@article{openai2024gpt4o,
  author    = "OpenAI",
  title     = "GPT-4o: Omnimodal Large Language Model",
  journal   = "OpenAI Technical Report",
  volume    = "",
  number    = "",
  pages     = "",
  year      = "2024"
}

@article{meta2024LLaMA3,
  author    = "Meta",
  title     = "The LLaMA 3 Herd of Models",
  journal   = "Meta AI Research Report",
  volume    = "",
  number    = "",
  pages     = "",
  year      = "2024"
}

@article{anthropic2024claude4sonnet,
  author    = "Anthropic",
  title     = "Claude 4 Sonnet: Advancements in Scalable Constitutional AI",
  journal   = "Anthropic Research Report",
  volume    = "",
  number    = "",
  pages     = "",
  year      = "2024"
}

@article{zerogpt2023,
  author    = "Rawad Baroud",
  title     = "ZeroGPT: AI Content Detection Tool for GPT-Generated Text",
  journal   = "ZeroGPT Research Blog",
  volume    = "",
  number    = "",
  pages     = "",
  url       = "https://www.zerogpt.com"
}

@article{originalityai2023,
  author    = "Jon Gillham",
  title     = "Originality.AI: Advanced AI Content Detection and Plagiarism Analysis Platform",
  journal   = "Originality.AI Research Blog",
  volume    = "",
  number    = "",
  pages     = "",
  url       = "https://originality.ai"
}

@article{mistral2024seba,
  author    = "Arthur Mensch",
  title     = "Mistral-Seba 24: Efficient and Open-Weight Large Language Model",
  journal   = "Mistral AI Research Report",
  volume    = "",
  number    = "",
  pages     = "",
  year      = "2024",
  url       = "https://mistral.ai/news/mistral-7b/"
}

@article{google2025gemma3,
  author    = "Google DeepMind",
  title     = "Gemma 3: Open Models for Responsible AI Innovation",
  journal   = "Google DeepMind Technical Report",
  volume    = "",
  number    = "",
  pages     = "",
  year      = "2025",
  url       = "https://deepmind.google/technologies/gemma/"
}

@article{deepseek2025v3_1,
  author    = "Deepseek",
  title     = "DeepSeek-V3.1: Advanced Multilingual Large Language Model",
  journal   = "DeepSeek AI Technical Report",
  volume    = "",
  number    = "",
  pages     = "",
  year      = "2025",
  url       = "https://www.deepseek.com/"
}

@article{alibaba2025qwen3,
  author    = "Alibaba",
  title     = "Qwen-3: Next-Generation Multilingual Foundation Model",
  journal   = "Alibaba Cloud Research Report",
  volume    = "",
  number    = "",
  pages     = "",
  year      = "2025",
  url       = "https://qwenlm.github.io/"
}

@inproceedings{Almohaimeed2023THOS,
  author    = "Almohaimeed, Saad",
  title     = "THOS: A Benchmark Dataset for Targeted Hate and Offensive Speech",
  booktitle = "Proceedings of the Data-centric Machine Learning Research Workshop at ICML 2023",
  year      = "2023",
  note      = "arXiv preprint arXiv:2311.06446"
}

@inproceedings{Almohaimeed_2024_ImplicitHate,
  author    = "Almohaimeed, Saad",
  title     = "Transfer learning and lexicon-based approaches for implicit hate speech detection: A comparative study of human and GPT-4 annotation",
  booktitle = "Proceedings of the 18th IEEE International Conference on Semantic Computing (ICSC-2024)",
  pages     = "142--147",
  year      = "2024"
}

@article{Raza2023NBIAS,
  author  = "Raza, Shaina and Garg, Muskan and Reji, Deepak John and Bashir, Syed Raza and Ding, Chen",
  title   = "NBIAS: A Natural Language Processing Framework for Bias Identification in Text",
  journal = "Expert Systems With Applications",
  volume  = "237",
  pages   = "121542",
  year    = "2023",
  doi     = "10.1016/j.eswa.2023.121542"
}

@inproceedings{Farrelly2023TopologicalBiasDetection,
  author    = "Farrelly, Colleen and Singh",
  title     = "Current Topological and Machine Learning Applications for Bias Detection in Text",
  booktitle = "2023 6th International Conference on Signal Processing and Information Security (ICSPIS)",
  pages     = "190--195",
  year      = "2023",
  doi       = "10.1109/ICSPIS60075.2023.10343824"
}

@article{Alshammari2024ArabicAIDetector,
  author  = "Alshammari, Hamed and El-Sayed, Ahmed and Elleithy, Khaled",
  title   = "AI-Generated Text Detector for Arabic Language Using Encoder-Based Transformer Architecture",
  journal = "Big Data and Cognitive Computing",
  volume  = "8",
  number  = "3",
  pages   = "32",
  year    = "2024",
  doi     = "10.3390/bdcc8030032"
}

@article{Chaka2024JALT,
  author	= "Chaka, Chaka",
  title		= "Reviewing the performance of {AI} detection tools in differentiating between {AI}-generated and human-written texts: A literature and integrative hybrid review",
  journal	= "Journal of Applied Learning \& Teaching",
  volume	= "7",
  number	= "1",
  pages		= "115--126",
  year		= "2024",
  doi		= "10.37074/jalt.2024.7.1.14"
}

@inproceedings{dugan2024raid,
  author    = "Dugan, Liam and Hwang, Alyssa and Trhlík, Filip and Ludan, Josh Magnus and Zhu, Andrew and Xu, Hainiu and Ippolito, Daphne and Callison-Burch, Chris",
  title     = "RAID: A Shared Benchmark for Robust Evaluation of Machine-Generated Text Detectors",
  booktitle = "Proceedings of the 62nd Annual Meeting of the Association for Computational Linguistics (ACL 2024)",
  pages     = "12463--12492",
  year      = "2024"
}

@inproceedings{wang2024m4,
  author    = "Wang, Yuxia and Mansurov, Jonibek and Ivanov, Petar and Su, Jinyan and Shelmanov, Artem and Tsvigun, Akim and Whitehouse, Chenxi and Mohamed Afzal, Osama and Mahmoud, Tarek and Sasaki, Toru and Arnold, Thomas and Aji, Alham Fikri and Habash, Nizar and Gurevych, Iryna and Nakov, Preslav",
  title     = "M4: Multi-generator, Multi-domain, and Multi-lingual Black-Box Machine-Generated Text Detection",
  booktitle = "Proceedings of the 18th Conference of the European Chapter of the Association for Computational Linguistics (Volume 1: Long Papers)",
  pages     = "1369--1407",
  year      = "2024",
  publisher = "Association for Computational Linguistics",
  url       = "https://aclanthology.org/2024.eacl-long.83",
  doi       = "10.18653/v1/2024.eacl-long.83"
}

@inproceedings{Li2024MAGE,
  author    = "Li, Yafu and Li, Qintong and Cui, Leyang and Bi, Wei and Wang, Zhilin and Wang, Longyue and Yang, Linyi and Shi, Shuming and Zhang, Yue",
  title     = "{MAGE:} Machine-Generated Text Detection in the Wild",
  booktitle = "Proceedings of the 62nd Annual Meeting of the Association for Computational Linguistics (ACL 2024)",
  pages     = "xxx-xxx",
  year      = "2024",
  url       = "https://aclanthology.org/2024.acl-long.3.pdf",
  note      = "Dataset available at https://github.com/yafuly/MAGE"
}

@inproceedings{Xu2024HCVar,
  author    = "Xu, Han and Ren, Jie and He, Pengfei and Zeng, Shenglai and Cui, Yingqian and Liu, Amy and Liu, Hui and Tang, Jiliang",
  title     = "{On the Generalization of Training-based ChatGPT Detection Methods}",
  booktitle = "Findings of the Association for Computational Linguistics: EMNLP 2024",
  pages     = "7223--7243",
  year      = "2024",
  publisher = "Association for Computational Linguistics",
  doi       = "10.18653/v1/2024.findings-emnlp.424",
  url       = "https://aclanthology.org/2024.findings-emnlp.424/"
}

@article{AuText2023,
  author  = "Sarvazyan, Areg Mikael and González, José Ángel and Franco-Salvador, Marc and Rangel, Francisco and Chulvi, Berta and Rosso, Paolo",
  title   = "{Overview of AuTexTification at IberLEF 2023: Detection and Attribution of Machine-Generated Text in Multiple Domains}",
  journal = "Procesamiento del Lenguaje Natural",
  volume  = "71",
  pages   = "275--288",
  year    = "2023"
}

@misc{HC3Plus,
  author       = "Zhenpeng Su",
  title        = "HC3 Plus: A Semantic-Invariant Human ChatGPT Comparison Corpus",
  howpublished = "Dataset release …",
  year         = "2024",
  note         = "Supports wide variety of languages"
}

@article{Gehrmann2019GLTR,
  author  = "Gehrmann, Sebastian and Strobelt, Hendrik and Rush, Alexander M.",
  title   = "{GLTR:} Statistical Detection and Visualization of Generated Text",
  journal = "arXiv preprint arXiv:1906.04043",
  year    = "2019"
}

@article{BERTDetect,
  author  = "Wang, Hao and Li, Jianwei and Li, Zhengyu",
  title   = "AI-Generated Text Detection and Classification Based on BERT Deep Learning Algorithm",
  journal = "arXiv preprint arXiv:2405.16422",
  year    = "2024"
}

@inproceedings{MULTITuDE,
  author    = "Macko, Dominik and Moro, Robert and Uchendu, Adaku and Lucas, Jason S. and Yamashita, Michiharu and Pikuliak, Matúš and Srba, Ivan and Le, Thai and Lee, Dongwon and Simko, Jakub and Bielikova, Maria",
  title     = "{MULTITuDE}: Large-Scale Multilingual Machine-Generated Text Detection Benchmark",
  booktitle = "Proceedings of the 2023 Conference on Empirical Methods in Natural Language Processing (EMNLP 2023)",
  year      = "2023"
}

@article{Al-Shaibani2025ArabicAIFingerprint,
  author  = "Al-Shaibani, Maged S. and Ahmed, Moataz",
  title   = "The Arabic AI Fingerprint: Stylometric Analysis and Detection of Large Language Models Text",
  journal = "arXiv preprint arXiv:2505.23276",
  year    = "2025"
}

@article{GPT-4,
  author  = "OpenAI",
  title   = "{GPT-4 Technical Report}",
  journal = "arXiv preprint arXiv:2303.08774",
  year    = "2023"
}

@article{Jais,
  author  = "Sengupta, Nandika and Sahu, Sudip Kumar and Jia, Bo and Katipomu, Sriram and Li, Hu and Koto, Felix and Marshall, Will and Gosal, Gem and Liu, Chen and Chen, Zhen and …",
  title   = "{Jais and Jais-Chat: Arabic-centric foundation and instruction-tuned open generative large language models}",
  journal = "arXiv preprint arXiv:2308.16149",
  year    = "2023"
}

@article{ALLaM,
  author  = "Bari, M. Saiful and Alnumay, Yazeed and Alzahrani, Norah A. and Alotaibi, Nouf M. and Alyahya, Hisham A. and AlRashed, Sultan and Mirza, Faisal A. and Alsubaie, Shaykhah Z. and Alahmed, Hassan A. and Alabduljabbar, Ghadah and …",
  title   = "{ALLaM:} Large Language Models for Arabic and English",
  journal = "arXiv preprint arXiv:2407.15390",
  year    = "2024"
}

@article{LLaMA3.1,
  author  = "Grattafiori, Aaron and Dubey, Abhimanyu and Jauhri, Abhinav and Pandey, Abhinav and Kadian, Abhishek and Al-Dahle, Ahmad and Letman, Aiesha and Mathur, Akhil and Schelten, Alan and Vaughan, Alex and Yang, Amy and Fan, Angela and Goyal, Anirudh and …",
  title   = "{The LLaMA 3 Herd of Models}",
  journal = "arXiv preprint arXiv:2407.21783",
  year    = "2024"
}

@inproceedings{XLM-R,
  author    = "Conneau, Alexis and Khandelwal, Kartikay and Goyal, Naman and Chaudhary, Vishrav and Wenzek, Guillaume and Guzmán, Francisco and Grave, Édouard and Ott, Myle and Zettlemoyer, Luke and Stoyanov, Veselin",
  title     = "{Unsupervised Cross-lingual Representation Learning at Scale}",
  booktitle = "Proceedings of the 58th Annual Meeting of the Association for Computational Linguistics (ACL 2020)",
  pages     = "8440--8451",
  year      = "2020",
  publisher = "Association for Computational Linguistics",
  doi       = "10.18653/v1/2020.acl-main.747"
}

@article{ANAD,
  author  = "Altamimi, Mohammed",
  title   = "{ANAD:} Arabic News Article Dataset",
  journal = "Data in Brief",
  volume  = "50",
  pages   = "109460",
  year    = "2023",
  doi     = "10.1016/j.dib.2023.109460"
}

@article{SANAD,
  author  = "Einea, Omar",
  title   = "{SANAD:} Single-Label Arabic News Articles Dataset for automatic text categorization",
  journal = "Data in Brief",
  volume  = "25",
  pages   = "104 076",
  year    = "2019",
  doi     = "10.1016/j.dib.2019.104076"
}

@misc{EASC,
  author       = "Melhaj, Mohammed",
  title        = "{EASC:} Essex Arabic Summaries Corpus",
  howpublished = "Corpus release, 153 articles + 765 summaries",
  year         = "2015",
  note         = "Available at SourceForge"
}

@online{site-ijssa,
  author       = "{IJSSA}",
  title        = "The Scientific Journal of Sport Science \& Arts",
  url          = "https://ijssa.journals.ekb.eg/journal/about?lang=en",
  note         = "Accessed 2025-11-08"
}

@online{site-WikipediaArabic,
  author       = "{Wikipedia}",
  title        = "Wikipedia: The Free Encyclopedia – A Collaborative Multilingual Knowledge Platform",
  url          = "https://ar.wikipedia.org",
  note         = "Accessed 2025-11-08"
}

@online{site-DollzterTravels2023,
  author       = "TheDollzter",
  title        = "The Dollzter Travels – A Travel and Lifestyle Blog Featuring Destinations, Experiences, and Guides Around the World",
  url          = "https://www.thedollztertravels.com/",
  note         = "Accessed 2025-11-08"
}

@online{site-UN2025,
  author       = "UN",
  title        = "United Nations – Official Website: Global Organization for Peace, Security, and Cooperation",
  url          = "https://www.un.org",
  note         = "Accessed 2025-11-08"
}

@online{site-Waqfeya2024,
  author       = "Waqfeya",
  title        = "Waqfeya - Arabic Digital Library for Free Downloadable Books and Manuscripts",
  url          = "https://waqfeya.com/",
  note         = "Accessed 2025-11-08"
}

@online{site-AAWSAT2025,
  author       = "Asharq Al-Awsat",
  title        = "Asharq Al-Awsat: The Leading Pan-Arab International Newspaper",
  year         = "2020",
  url          = "https://aawsat.com",
  note         = "Accessed 2025-11-08"
}

@online{site-Coins4Arab2025,
  author       = "Coins4Arab",
  title        = "Coins4Arab – Arabic Cryptocurrency Forum",
  year         = "2020",
  url          = "https://coins4arab.com/vb/index.php",
  note         = "Accessed 2025-11-08"
}

@article{GPT35Turbo,
  author  = "OpenAI",
  title   = "{GPT-3.5-Turbo: Advancements in Instruction-Tuned Large Language Models}",
  journal = "OpenAI Technical Report",
  year    = "2023",
  url     = "https://platform.openai.com/docs/models/gpt-3-5-turbo",
  note    = "Accessed 2025-11-08"
}

@misc{Meta-LLaMA4-17bMaverick,
  author       = "{Meta}",
  title        = "{LLaMA-4 Maverick-17B-128E-Instruct-FP8}",
  howpublished = "\url{https://huggingface.co/meta-LLaMA/LLaMA-4-Maverick-17B-128E-Instruct-FP8}",
  year         = "2025",
  note         = "Mixture-of-Experts (MoE) model: 17B active params / 128 experts / FP8 quantization. Released April 2025. Knowledge cutoff Aug 2024.  :contentReference[oaicite:2]{index=2}"
}

@misc{KimiK2Instruct,
  author       = "Moonshot",
  title        = "{Kimi-K2-Instruct: A 1 Trillion-Parameter Mixture-of-Experts Language Model (32 B active) for Agentic Intelligence}",
  howpublished = "\url{https://huggingface.co/moonshotai/Kimi-K2-Instruct}",
  year         = "2025",
  note         = "Released July 2025; supports 128 K context window.  :contentReference[oaicite:2]{index=2}"
}

@online{SmodinAIDetector2025,
  author       = "Kevin",
  title        = "Smodin AI Detector – Accurate AI Checker for ChatGPT, GPT-5 \& Gemini",
  url          = "https://smodin.io/ai-content-detector",
  note         = "Accessed 2025-11-09"
}

@online{IsgenAIDetector,
  author       = "{Isgen.ai}",
  title        = "Isgen AI Detector – Detect AI-Generated Text (Arabic \& multilingual)",
  url          = "https://isgen.ai/ar",
  note         = "Accessed 2025-11-09"
}

@inproceedings{SentenceBERT,
  author    = "Reimers, Nils and Gurevych, Iryna",
  title     = "Making Monolingual Sentence Embeddings Multilingual using Knowledge Distillation",
  booktitle = "Proceedings of the 2020 Conference on Empirical Methods in Natural Language Processing (EMNLP)",
  pages     = "in press",
  year      = "2020"
}

@inproceedings{MPNet,
  author    = "Song, Kaitao and Tan, Xu and Qin, Tao and Lu, Jianfeng and Liu, Tie-Yan",
  title     = "{MPNet:} Masked and Permuted Pre-training for Language Understanding",
  booktitle = "Proceedings of the 34th Conference on Neural Information Processing Systems (NeurIPS 2020)",
  pages     = "16857--16867",
  year      = "2020",
  publisher = "Curran Associates, Inc.",
  doi       = "10.5555/3495724.3497138"
}

@inproceedings{Almohaimeed2024GAT-SQL,
  author    = "Almohaimeed, Saleh",
  title     = "{GAT-SQL:} An Advanced Prompt Engineering Approach for Effective Text-to-SQL Interactions",
  booktitle = "2024 IEEE Congress on Evolutionary Computation (CEC)",
  pages     = "xxx--xxx",
  year      = "2024",
  doi       = "10.1109/CEC60901.2024.xxxxxx"
}

@online{OpenAIPlatform,
  author       = "OpenAI",
  title        = "OpenAI Platform – Developer API \& resources",
  url          = "https://platform.openai.com/",
  note         = "Accessed 2025-11-09"
}
\FloatBarrier
\newpage

\begin{appendices}

\section{Numerical representation of figures 2 to 7}\label{AP1}

\begin{table}[h]
\centering
\caption{AI-text detection accuracy of 6 LLMs, where columns refer to those used for detection, and rows refer to those used for polishing the text under 10\% polishing. ORG refers to the accuracy of the detector under no polishing.}
\label{10P-polishing}
\begin{tabular}{lcccccc}
\toprule
\textbf{Models} & \textbf{GPT-4o} & \textbf{Deepseek 3.1} & \textbf{Mistral} & \textbf{Kimi K2} & \textbf{LLAMA-4 17B} & \textbf{Claude-4 sonnet} \\
\midrule
ORG     & 99.2 & 96.87 & 89.76 & 74.2 & 74.61 & 83.51 \\
GPT-4o    & 99.2 & 97.12 & 98.17 & 62.3 & 74.08 & 73.99 \\
GPT-3.5    & 97.88 & 90.7 & 97.88 & 59.15 & 69.58 & 72.54 \\
LLAMA-3 70B   & 95.26 & 87.63 & 95.53 & 60.5 & 79.73 & 57.63 \\
Deepseek 3.1   & 99.48 & 96.6 & 98.43 & 72.97 & 77.35 & 75.13 \\
Qwen-3    & 99.7 & 97.23 & 99.38 & 78.1 & 86.67 & 76.31 \\
Mistral    & 97.67 & 96.63 & 97.15 & 69.17 & 72.54 & 65.03 \\
LLAMA-4 17B  & 98.69 & 94.5 & 95.55 & 70.42 & 74.08 & 73.04 \\
GEMMA   & 98.22 & 95.58 & 96.89 & 71.43 & 75.58 & 65.71 \\
Claude-4 sonnet & 99.48 & 94.76 & 95.81 & 73.3 & 75.13 & 80.89 \\
Kimi K2    & 98.43 & 94.79 & 96.35 & 71.61 & 80.73 & 76.04 \\
\bottomrule
\end{tabular}
\end{table}

\begin{table}[h]
\centering
\caption{AI-text detection accuracy of 6 LLMs, where columns refer to those used for detection, and rows refer to those used for polishing the text under 25\% polishing. ORG refers to the accuracy of the detector under no polishing.}
\label{25P-polishing}
\begin{tabular}{lcccccc}
\toprule
\textbf{Detectors} & \textbf{GPT-4o} & \textbf{Deepseek 3.1} & \textbf{Mistral} & \textbf{Kimi K2} & \textbf{LLAMA-4 17B} & \textbf{Claude-4 sonnet} \\
\midrule
ORG     & 99.2 & 96.87 & 89.76 & 74.2 & 74.61 & 83.51 \\
GPT-4o    & 99.2 & 95.04 & 97.90 & 70.76 & 73.80 & 73.37 \\
GPT-3.5    & 98.21 & 94.97 & 97.88 & 66.93 & 70.31 & 73.99 \\
LLAMA-3 70B   & 87.04 & 71.69 & 75.75 & 49.7 & 69.05 & 35.98 \\
Deepseek 3.1   & 99.29 & 96.72 & 98.70 & 71.73 & 76.88 & 72.13 \\
Qwen-3    & 99.32 & 97.10 & 97.14 & 72.80 & 84.22 & 74.92 \\
Mistral    & 97.43 & 93.00 & 97.22 & 64.61 & 70.92 & 60.77 \\
LLAMA-4 17B  & 96.30 & 92.00 & 91.87 & 65.18 & 74.42 & 70.42 \\
GEMMA   & 98.30 & 96.10 & 96.80 & 66.84 & 71.30 & 63.51 \\
Claude-4 sonnet & 99.06 & 93.67 & 94.71 & 70.08 & 75.01 & 79.58 \\
Kimi K2    & 98.22 & 92.91 & 93.63 & 62.30 & 81.62 & 63.35 \\
\bottomrule
\end{tabular}
\end{table}

\begin{table}[h]
\centering
\caption{AI-text detection accuracy of 6 LLMs, where columns refer to those used for detection, and rows refer to those used for polishing the text under 50\% polishing. ORG refers to the accuracy of the detector under no polishing.}
\label{50P-polishing}
\begin{tabular}{lcccccc}
\toprule
\textbf{Detectors} & \textbf{GPT-4o} & \textbf{Deepseek 3.1} & \textbf{Mistral} & \textbf{Kimi K2} & \textbf{LLAMA-4 17B} & \textbf{Claude-4 sonnet} \\
\midrule
ORG     & 99.2 & 96.87 & 89.76 & 74.2 & 74.61 & 83.51 \\
GPT-4o    & 98.43 & 95.03 & 97.01 & 73.28 & 72.41 & 71.47 \\
GPT-3.5    & 98.22 & 93.42 & 96.89 & 58.05 & 71.68 & 72.05 \\
LLAMA-3 70B   & 83.64 & 67.02 & 71.96 & 35.4 & 79.9 & 30.08 \\
Deepseek 3.1   & 98.91 & 96.42 & 98.12 & 62.57 & 76.12 & 69.49 \\
Qwen-3    & 99.64 & 96.99 & 99.83 & 74.2 & 84.70 & 75.31 \\
Mistral    & 97.62 & 95.06 & 97.45 & 61.98 & 67.95 & 57.00 \\
LLAMA-4 17B  & 95.10 & 89.23 & 87.12 & 66.23 & 73.04 & 66.23 \\
GEMMA   & 97.90 & 95.33 & 96.12 & 57.81 & 73.25 & 60.90 \\
Claude-4 sonnet & 98.55 & 91.44 & 91.10 & 59.69 & 74.13 & 68.59 \\
Kimi K2    & 99.12 & 92.88 & 92.98 & 58.68 & 80.33 & 70.26 \\
\bottomrule
\end{tabular}
\end{table}

\begin{table}[h]
\centering
\caption{AI-text detection accuracy of 6 LLMs, where columns refer to those used for detection, and rows refer to those used for polishing the text under 75\% polishing. ORG refers to the accuracy of the detector under no polishing.}
\label{75P-polishing}
\begin{tabular}{lcccccc}
\toprule
\textbf{Detectors} & \textbf{GPT-4o} & \textbf{Deepseek 3.1} & \textbf{Mistral} & \textbf{Kimi K2} & \textbf{LLAMA-4 17B} & \textbf{Claude-4 sonnet} \\
\midrule
ORG     & 99.2 & 96.87 & 89.76 & 74.2 & 74.61 & 83.51 \\
GPT-4o    & 97.6 & 95.29 & 96.86 & 67.8 & 72.51 & 62.83 \\
GPT-3.5    & 98.4 & 96.02 & 97.08 & 65.25 & 74.54 & 72.49 \\
LLAMA-3 70B   & 61.5 & 50.67 & 69.6 & 17.3 & 37.33 & 12.53 \\
Deepseek 3.1   & 98.95 & 96.33 & 98.43 & 64.04 & 76.12 & 68.24 \\
Qwen-3    & 98.84 & 96.81 & 97.39 & 67.54 & 85.10 & 71.88 \\
Mistral    & 97.13 & 93.47 & 96.34 & 57.85 & 66.58 & 53.00 \\
LLAMA-4 17B  & 95.26 & 88.45 & 86.88 & 52.37 & 72.20 & 44.36 \\
GEMMA   & 98.17 & 96.03 & 95.56 & 67.62 & 72.85 & 60.57 \\
Claude-4 sonnet & 98.17 & 90.84 & 92.67 & 67.54 & 74.87 & 63.35 \\
Kimi K2    & 99.21 & 92.13 & 92.65 & 66.67 & 82.67 & 67.19 \\
\bottomrule
\end{tabular}
\end{table}

\FloatBarrier

\newpage
\section{Examples of System and User Prompts}\label{AP2}

\begin{figure}[!ht]
\vfill
\begin{center}
\includegraphics[width=\textwidth]{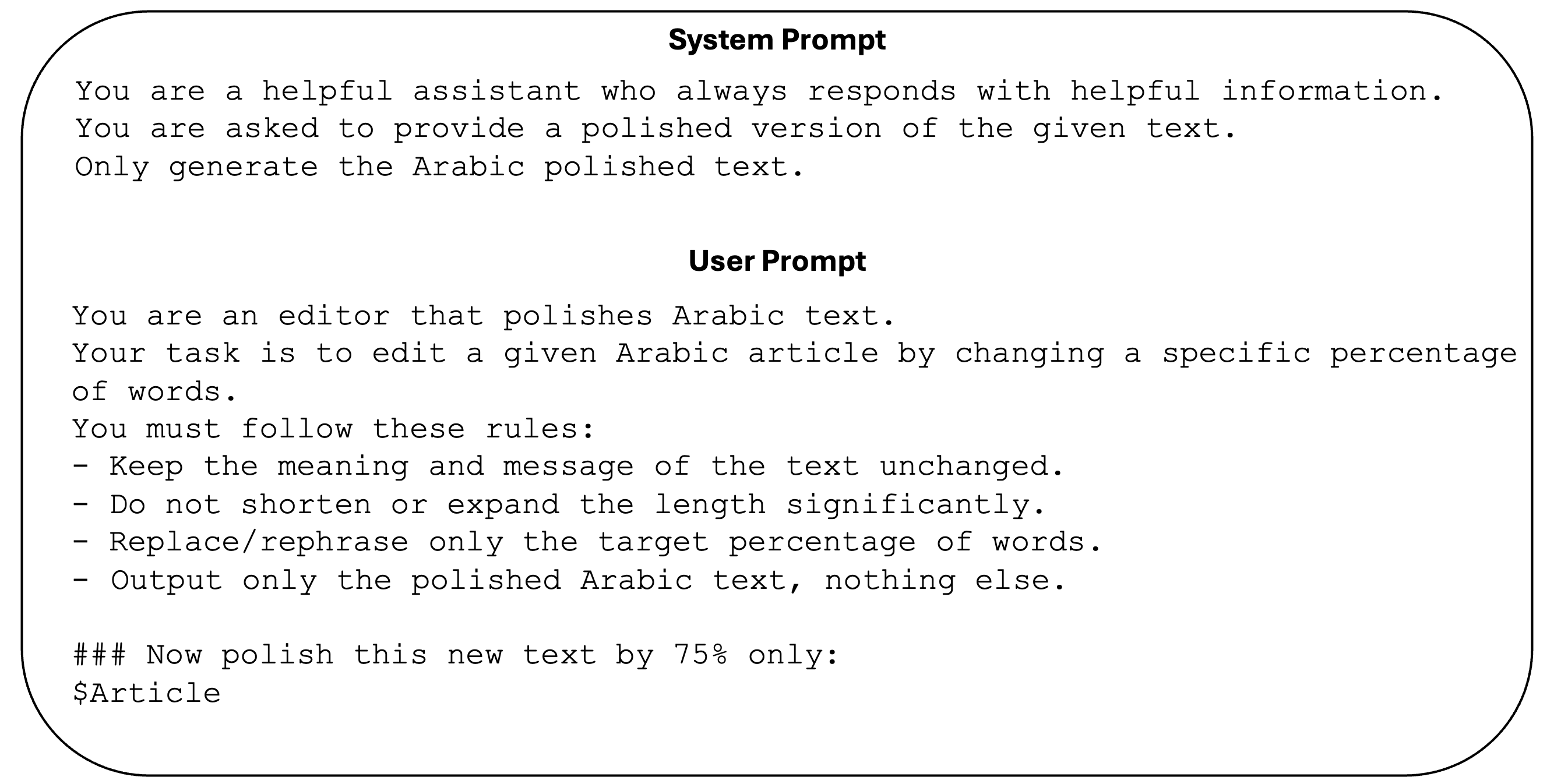}
\caption{This is an example of the prompt instruction used by the 10 LLMs to polish the human-written articles, where the percentage specified will be one of four: 10\%, 25\%, 50\% and 75\%.}
\label{polish prompt}
\end{center}
\vfill
\end{figure}

\begin{figure}[!ht]
\vfill
\begin{center}
\includegraphics[width=1\textwidth]{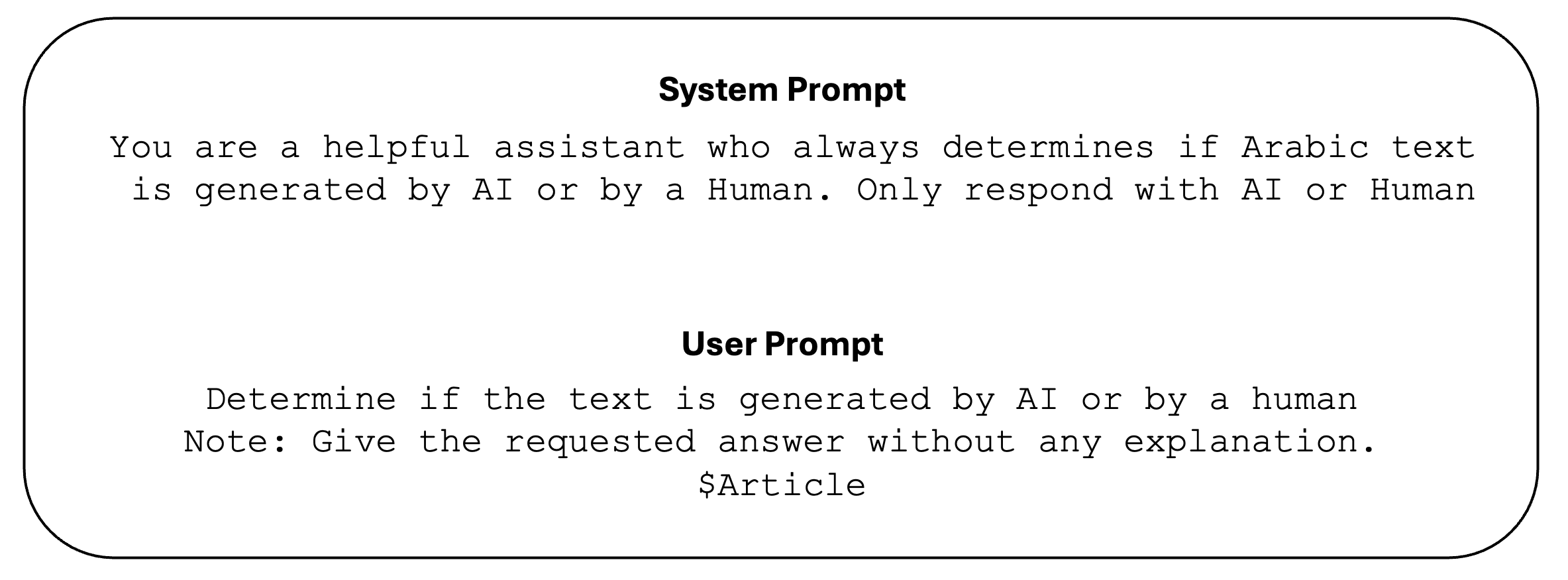}
\caption{This is an example of the prompt instruction used by the 10 LLMs to detect AI-generated articles. we emphasize that LLM should respond by human or artificial intelligence, as we see that this emphasis will increase their compliance with this instruction}
\label{Detection prompt}
\end{center}
\vfill
\end{figure}

\end{appendices}

\end{document}